\documentclass[letterpaper, 10 pt, conference]{ieeeconf}  

\IEEEoverridecommandlockouts                              
\usepackage{booktabs}
\usepackage{graphicx}
\usepackage{amsmath}
\usepackage{amssymb}
\usepackage{graphicx}
\usepackage{color}
\usepackage{float}
\usepackage{units}
\usepackage{footnote}
\makesavenoteenv{tabular}
\usepackage{pbox}
\makesavenoteenv{table}
\usepackage[all=normal,paragraphs=tight,floats=tight,mathspacing=tight,wordspacing=tight,charwidths=normal,mathdisplays=normal,leading=normal]{savetrees}
\DeclareMathOperator*{\argmin}{\arg\min}

\newcommand{\camposetime}[1]{^{c_{#1}}\mathbf{T}_w}
\newcommand{\campose}[0]{^c\mathbf{T}_w}

\newcommand{\pointidx}[2]{^{#1}\mathbf{X}_{#2}}
\newcommand{\point}[1]{^{#1}\mathbf{X}}
\newcommand{\imgpoint}[1]{\mathbf{x}_{#1}}
\newcommand{\imgpointij}[2]{\mathbf{x}_{#1, #2}}

\title{\LARGE \bf
S$^3$LAM: Structured Scene SLAM 
}

\author{Mathieu Gonzalez$^{1}$, Eric Marchand$^{2}$, Amine Kacete$^{1}$ and Jerome Royan$^{1}$
\thanks{$^{1}$ Mathieu Gonzalez, Amine Kacete and Jerome Royan are with the Institute of Research and Technology b$<>$com
       {\tt\small \{mathieu.gonzalez,amine.kacete, jerome.roy\}@b-com.com}}%
\thanks{$^{2}$ Eric Marchand is with Univ Rennes, Inria, IRISA, CNRS, Rennes, France,
        {\tt\small Eric.Marchand@irisa.fr}}%
}

\begin{document}

\maketitle
\thispagestyle{empty}
\pagestyle{empty}

\begin{abstract}

We propose a new SLAM system that uses the semantic segmentation of objects and structures in the scene. Semantic information is relevant as it contains high level information which may make SLAM more accurate and robust. Our contribution is twofold: i) A new SLAM system based on ORB-SLAM2 that creates a semantic map made of clusters of points corresponding to objects instances and structures in the scene. ii) A modification of the classical Bundle Adjustment formulation to constrain each cluster using geometrical priors, which improves both camera localization and reconstruction and enables a better understanding of the scene. We evaluate our approach on sequences from several public datasets and show that it improves camera pose estimation with respect to state of the art.
\end{abstract}
~\\

\section{INTRODUCTION}

\label{sec:intro}
The goal of SLAM is to construct a map of the environment seen by a moving camera while simultaneously estimating the pose of the camera. It is a fundamental algorithm for robotics and augmented reality (AR) that has seen many improvements during the past few years and is now able to correctly estimate the pose of a camera in small and large scale scenes \cite{mur2017orb, engel2014lsd}. The map can take different forms, sparse \cite{mur2017orb}, semi-dense \cite{engel2014lsd} but is always purely geometric and lacks semantic meaning which is an important information to make SLAM more robust and accurate \cite{cadena2016past, rosinol2020kimera}. Furthermore the lack of semantic information may limit the applications that can make use of this map. For example mobile robots may need to recognize objects in the scene for path planing. AR applications can as well use semantic information to overlay contextual information on specific parts of the image and virtually interact with some objects \cite{runz2018maskfusion}.
With the rise of CNNs, methods for object detection \cite{redmon2016you} and segmentation \cite{he2017mask, kirillov2019panoptic} are now available for practical use applications. Several studies have been conducted to merge those methods with SLAM by fusing multiple segmentations to obtain consistent semantic maps \cite{tateno2017cnn,mccormac2017semanticfusion}. Furthermore a semantic comprehension of the environment may improve SLAM itself at several stages of the pipeline, more particularly for long term re-localization \cite{schonberger2018semantic, toft2017long, toft2018semantic} and dynamic objects handling \cite{bescos2018dynaslam, runz2018maskfusion, huang2020clustervo}. Some works use objects as high level landmarks \cite{yang2019cubeslam, salas2013slam++, hosseinzadeh2019real, nicholson2018quadricslam, galvez2016real}. To do so, they can make use of specific objects \cite{salas2013slam++, galvez2016real, fioraio2013joint, civera2011towards} at the disadvantage of requiring a specialised object pose estimation algorithm \cite{rad2017bb8, gonzalez2021l6dnet, xiang2017posecnn}. To discard this constraint some works propose to represent objects in a generic way using for example quadrics \cite{hosseinzadeh2019real, nicholson2018quadricslam} or 3D bounding boxes \cite{yang2019cubeslam} and use a generic object detector. Planes can be seen as specific objects that are numerous in many environments and can be also integrated into SLAM \cite{hosseinzadeh2019real, kaess2015simultaneous, ardnt2020from, yang2016pop}.    

\begin{figure}
    \centering
    \includegraphics[width=\columnwidth]{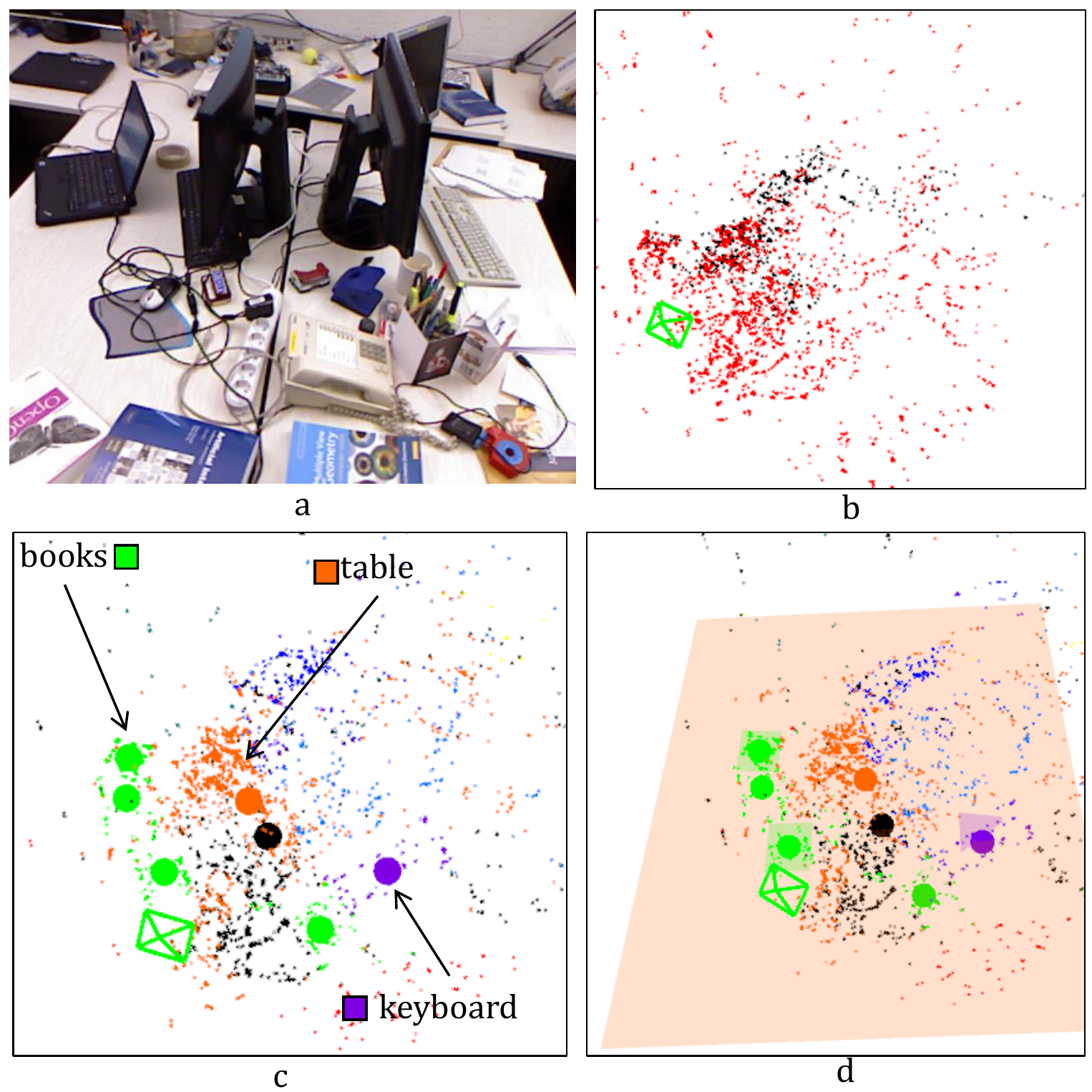}
    \caption{(a) frame from the sequence fr1\_desk. Comparison of (b) a map built by ORB-SLAM2, (c) a map of clusters where each cluster centroid is represented with a big sphere (books are in green, table in orange, keyboard in purple, other objects are not used) and (d) a map of clusters with estimated planes. }
    \label{fig:first_fig}
\end{figure}

In this paper we present a monocular SLAM system, called S$^3$LAM for Structured Scene SLAM, based on the state-of-the-art ORB-SLAM2 \cite{mur2017orb} that can segment generic objects in the scene using a panoptic segmentation CNN, namely detectron2 \cite{wu2019detectron2}. 
We propose to create a new scene representation in which objects are seen as clusters of triangulated 3D points with semantic information. This allows us to create a semantic map with object instances and structures as shown in figure \ref{fig:first_fig}. Using this map of clusters we make use of prior information about object classes to constrain the map, which improves camera localization, and provides a higher level semantic map as shown in figure \ref{fig:first_fig}.

~\
In summary contributions presented in this paper are:
\begin{itemize}
\item A SLAM framework that can detect object instances in the scene to create clusters of 3D points corresponding to such objects.
    
    \item A monocular SLAM system that can infer structures from clusters and constrain the map given such estimations.
    
    \item An evaluation of our approach on sequences from several public datasets which demonstrates the benefits of our method in term of camera pose precision.
\end{itemize}

The rest of the paper is described as follows. First we describe related work on classical and semantic SLAM. Then we describe our approach to create a structured map and make use of those structures. Finally we demonstrate the benefits of our approach on several sequences from a public dataset.
\section{Related work: Semantic SLAM}
Classically SLAM is solved by estimating the pose of the camera and the map of the environment by maximizing the likelihood of those variables given image measurements. ORB-SLAM2 \cite{mur2017orb} is considered to be the  current State-Of-The-Art of visual SLAM. In their system the measurements are sparse ORB \cite{rublee2011orb} keypoints extracted from images. The system is divided in threads which estimate the pose of the camera and the 3D position of map points. This strategy inspired from \cite{mouragnon2006real, klein2007parallel} allows to refine the estimations with a local Bundle Adjustment \cite{triggs1999bundle} which minimizes the reprojection error of map points. LSD-SLAM \cite{engel2014lsd} directly uses pixels on high gradient regions of the image to estimate the pose of the camera and build a semi-dense map. The pose of the keyframes are as well refined using pose graph optimization. 

Semantic information can be calculated using CNNs and then injected into a SLAM map  \cite{mccormac2017semanticfusion, sunderhauf2017meaningful, tateno2017cnn}. SemanticFusion \cite{mccormac2017semanticfusion} is a dense SLAM which computes a pixelwise probability distribution for each frame and fuse the results for each surfel using a bayesian approach, which gives a semantic dense map. The semantic map can then be used to improve the SLAM.

Object based SLAM systems consist in detecting objects in the scene and inserting them in the map to add constraints between frames, thus adding temporal consistency \cite{salas2013slam++, fioraio2013joint, galvez2016real, yang2019cubeslam, hosseinzadeh2019real, nicholson2018quadricslam}. This can bring robustness to the SLAM and accuracy by having access to the objects scale. Those systems can be divided into two main categories.
The first one consist of object based SLAM systems that use specific objects \cite{salas2013slam++, galvez2016real, fioraio2013joint, civera2011towards}. SLAM++ \cite{salas2013slam++} proposes to estimate the 6 DoF pose of objects in the scene from RGB-D images. Each estimated object is rendered using its mesh and the pose of the camera is estimated by minimizing the ICP error with the live depth frame. SLAM++ builds a graph of keyframes and objects as a map and optimizes the pose graph. On the other hand some works propose to model objects using quadrics \cite{hosseinzadeh2019real, nicholson2018quadricslam} or 3D bounding boxes \cite{yang2019cubeslam}. QuadricSLAM \cite{nicholson2018quadricslam} uses quadrics for localization and mapping. The main idea is to generate quadrics from 2D bounding boxes predicted by an object detection network. The quadrics parameters and keyframe poses are then refined in a BA so that the 2D projection of quadrics tightly fits the 2D bounding boxes.

Planar SLAM systems consist in detecting planar structures in the scene and using them as high level landmarks \cite{kaess2015simultaneous, ardnt2020from, hosseinzadeh2019real, yang2019monocular, yang2016pop, hsiao2017keyframe}. The goal is threefold: first, planes are usually large structures and can thus constrain different parts of a scene without visual overlap. Second, some man-made planar structures do not contain valuable information for keypoint based SLAM, for example white walls, hence using planes as landmarks can enable tracking and mapping in those challenging cases. Finally detecting planes in the scene allows to get a better understanding of its physical structure which can enable interaction, contrary to a simple sparse point cloud.
\cite{kaess2015simultaneous} proposes to use only planes as landmarks. Planes are extracted from RGB-D images and injected in a graph based SLAM using a minimal representation which allows them to be optimized with keyframes trajectory.
\cite{hosseinzadeh2019real} uses planes to constrain the SLAM map. Planes are estimated from 3 different neural networks that estimate depth, normals and plane segmentation. From these redundant estimations planes are inserted in the map.  The point-plane distance is then minimized within the BA.
\cite{ardnt2020from} proposes a monocular SLAM using planes to constrain the structure of the scene. Planes are estimated using a RANSAC on the whole map, thus they need to be large enough and the framerate need to be low enough for the RANSAC to find the panes. In-plane points are then projected onto the estimated planes and both 2D plane points as well as plane parameters are optimized to minimize the reprojection error. 

\section{S$^3$LAM: A cluster based SLAM }
\begin{figure*}[ht]
    \centering
    \includegraphics[scale=0.48]{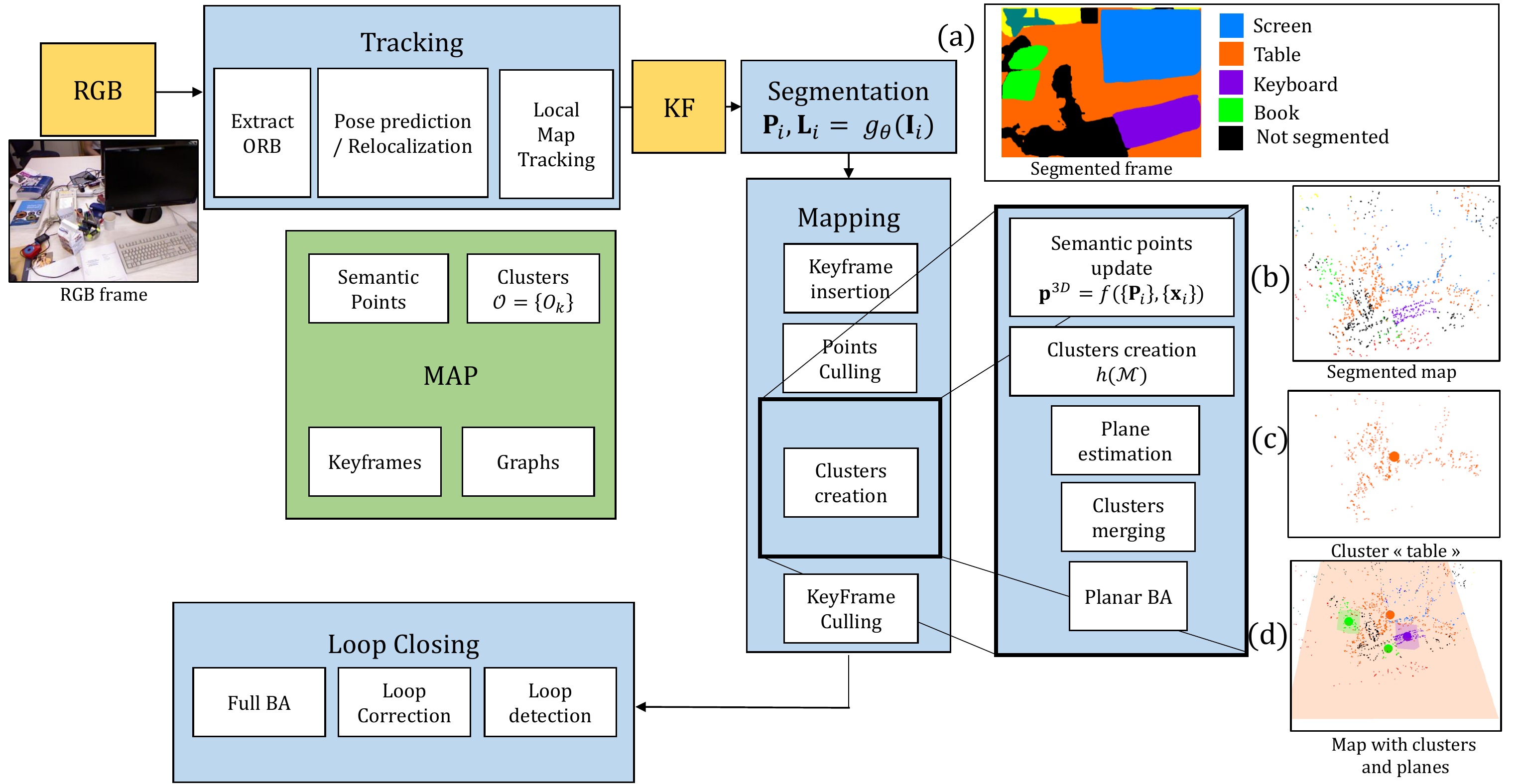}
    \caption{Illustration of our pipeline integrated in \cite{mur2017orb}. (a) A CNN segments the keyframes, (b) the output of the CNN allows us to compute the probability distribution of map points, (c) we create clusters of points based on their semantic class and fit planes for planar classes, (d) we apply our BA constrained by planes.}
    \label{fig:main_fig}
\end{figure*}

In S$^3$LAM the map is represented as a set of point clouds, grouped according to the object instance they belong. Our goal is to use prior knowledge about these objects to enrich the SLAM, improve camera pose estimation as well as to obtain a better representation of the structure of the scene.
Our goal is to estimate the pose of a monocular camera moving in 3D space, represented at time $i$ by the 6 DoF transformation between world frame $\mathcal{F}_w$ and camera frame $\mathcal{F}_{c_i}$ with the homogeneous matrix that defines the transformation denoted $\camposetime{i} \in SE(3)$.
The pipeline of our approach can be seen in figure \ref{fig:main_fig}. We segment each keyframe using a panoptic segmentation network. In the mapping thread we update the class distribution of map points using the output of the panoptic segmentation network. Segmented map points allow us to create semantic clusters that uniquely correspond to object instances. For some clusters of classes corresponding to planar objects a plane is fitted using the 3D points and a merging step avoids the creation of duplicated clusters. A bundle adjustment with a planar constraint is then applied to refine camera pose estimation and map points position.
Compared to state of the art our approach relates to object based methods that represent objects as quadrics \cite{nicholson2018quadricslam, hosseinzadeh2018structure, hosseinzadeh2019real} that are very generic and approaches that estimate planes to represent the map \cite{ardnt2020from, hosseinzadeh2018structure, hosseinzadeh2019real}. However contrary to approaches that use quadrics, ours yields a representation that is closer to reality, which can further improve camera pose estimation accuracy. And contrary to approaches that use planes, ours does not need depth information, nor specific CNN to estimate planes, nor very large planes and can work in near real time.

\subsection{Clusters Creation.}
In this section we show how we manage to cluster points, according to the object instance they belong using semantic information. 

~\\
\textbf{Panoptic segmentation:}
Unlike most recent works \cite{hosseinzadeh2019real, yang2019cubeslam} we do not consider object bounding boxes as input but rather the panoptic segmentation of the image. Panoptic segmentation is a combination of semantic segmentation where each pixel is classified in a given class and instance segmentation where multiple objects of the same class are segmented separately. While panoptic segmentation is harder and takes longer to obtain it allows us to naturally know which keypoints belong to detected objects. Indeed while bounding boxes give the coarse location and size of an object in the image they do not separate the object from the background within the box. Hence a refinement process is required \cite{huang2020clustervo}. Moreover contrary to object detection CNN, panoptic segmentation networks, like semantic segmentation networks, are not limited to objects and can segment whole areas in the image, such as floor,  which correspond to the global structure of the scene. However contrary to semantic segmentation networks, panoptic segmentation separates multiple instances of a single class, allowing to treat each object separately.

The downside of the additional information brought by panoptic segmentation is its complexity, while object detectors can easily process tens of images per second \cite{redmon2016you}, most recent panoptic segmentation networks run only at 10 to 20 fps.
However we do not need to segment images at frame rate. As shown in \cite{mccormac2017semanticfusion}, segmenting frames with a low frequency leads to a small drop in mapping segmentation accuracy while allowing the SLAM to run in real time. Moreover panoptic segmentation is an active field of research and real time networks can be expected in the near future.

~\\
\indent To represent the panoptic segmentation network we define a function $g_\theta(\mathbf{I}_i) \rightarrow \mathbf{P}_i, \mathbf{L}_i $ which, given an RGB image at time $i$, $\mathbf{I}_i$ and $\theta$, the network parameters, yields a probability map $\mathbf{P}_i \in [0,1]^{W \times H \times C}$\footnote{$W$ and $H$ represent the width and height of an RGB image, $C$ is the total number of semantic classes.}. Thus for each pixel $\imgpoint{}=(u,v)$ in the image we can obtain a probability distribution $(\mathbf{P}_i(u,v,1), ..., \mathbf{P}_i(u,v,C))$ where $\mathbf{P}_i(u,v,c)$ corresponds to the probability that this pixel belongs to class $c$. The second output of the network is the instance map $\mathbf{L}_i \in \mathbb{N}^{W \times H}$ in which each object is segmented and given a unique id.

One problem of panoptic segmentation however is that the instance map is not temporally consistent, meaning that we can not simply rely on the object ids to create the map. To solve this problem we propose a 2 stages strategy: at the level of the segmentation network and in the SLAM. First, for each detected instance in $\mathcal{L}^i$ we compute the IOU with all instances of the same class in $\mathcal{L}^{i-1}$ and $\mathcal{L}^{i-2}$ and take its maximum to track the id. We consider an instance to be well tracked if the IOU is above a threshold $\tau_{i-1} = 0.65$ for $\mathcal{L}^{i-1}$ and $\tau_{i-2} = 0.4$ for $\mathcal{L}^{i-2}$. A similar approach can also be applied by propagating the 2D id using optical flow, which requires more computations. Using our strategy, clusters are well defined. However when a cluster that has left the camera field of view reappears in the image, the segmentation network creates a new instance, which leads to the creation of a new cluster. To avoid creating infinitely many clusters and propagate the cluster id we define a merging function to fuse clusters together when the distance between their centroid is lower than a threshold $\tau_{merge}$ and more than 80\% of clusters points descriptors match. This approach can also be robustified by reprojecting cluster points in 2D and assigning them the id of the segment in which they fall. The combination of those approaches in 2D and 3D allows us to create a sparse consistent map.

~\\
\textbf{Clusters creation:}
Following the 2D segmentation, we define a function $f(\{\mathbf{P}_i\}, \{\imgpoint{i}\}) \rightarrow  \mathbf{p}^{3D} $ where $\{\mathbf{P}_{i}\}$ is a set of probability maps at different times, $\{\imgpoint{i}\}$  is a set of keypoints corresponding to one 3D point $\point{w}$ and $\mathbf{p}^{3D} = (p_1,..., p_C)$ is its probability distribution. This function is the fusion of multiple observations and can be written using Bayes rule, which was inspired by \cite{mccormac2017semanticfusion} that we adapt to work with our sparse monocular approach as it is mostly used by dense SLAM systems using RGB-D inputs.
\begin{equation}
\label{eq:bayes_update}
p_c = \mathbb{P}(c | \{\mathbf{P}_i\}) = \frac{1}{Z} \mathbb{P}(c | \{\mathbf{P}_{i-1}\}) \mathbf{P}_i(u,v,c)
\end{equation}
where $c$ is the class label of $\point{w}$ and $Z$ is a normalization factor.
Hence $f$ allows us to obtain a semantic map $\mathcal{M} = \{(\point{w}, \mathbf{p}^{3D}, c^*, l)_j\}$, where each point $\point{w}$ has a probability distribution  $\mathbf{p}^{3D}$, an id $l$ extracted from the instance map $\mathbf{L}^{i}$ as well as a semantic class $c^* = \underset{c}{\mathrm{argmax}}{ \; \mathbf{p}^{3D}}$. 
This fusion allows the map to be temporally consistent, even if the panoptic segmentation is noisy.
~\

Using this semantic map we can define a clustering function $h(\mathcal{M}) \rightarrow \mathcal{O}$ where $\mathcal{O}$ is a partition of $\mathcal{M}$ in $K$ clusters $\mathcal{O} = \{O_{k, k \in [1,K]}\}$. This function groups points according to their semantic class and instance.
Each cluster can be defined as $O_k = \{\{\point{w}\}, c_k\}$ where $\{\point{w}\}$ is the position of a set of points belonging to the cluster and $c_k$ is the cluster class.

\subsection{Map Optimization from structure estimation.}
~\\
\textbf{Structure estimation:} Most man-made objects can be approximated with a more or less complex geometrical model, from a simple plane or box to the exact 3D model of the object. The advantage of approximating objects is twofold. First, we can add constraints to the optimization process to improve pose estimation. Second, we obtain a more physically accurate representation of the world by understanding the structures within it, contrary to recent methods that use quadrics to represent objects, which only roughly represent the spatial extent of objects but not their shape \cite{hosseinzadeh2019real, nicholson2018quadricslam}.
In our work as an example we propose to model some objects using planes. Not only do we model large surfaces such as tables, walls or floor but we also model small objects like keyboards and books. While this hypothesis is more restrictive than using quadrics we argue that it stays highly generic as we estimated that about 25\% of classes from the COCO dataset can be represented with planes. 
Planes are represented using the classical 4D vector $\pi = (a,b,c,d)^\top$ with $||\pi||_2 = 1$ \cite{kaess2015simultaneous} and planar points $\mathbf{X}$ in homogeneous coordinates satisfy the following equation:
\begin{equation}
    \pi^\top\mathbf{X} = 0.
\end{equation}

Contrary to most planar SLAM systems we do not need to use multiple specific CNNs \cite{hosseinzadeh2019real} or depth \cite{kaess2015simultaneous} to estimate plane parameters, which limit the applicability of those systems. Instead for each a priori planar cluster, we fit a plane using the triangulated 3D points of this cluster in the world coordinates system. This is done using an SVD inside a RANSAC loop to make the estimation more robust to wrong classification and triangulation similarly to \cite{ardnt2020from}. However contrary to \cite{ardnt2020from} we are not limited to simple scenes with few very large planes. Indeed as we create semantic clusters we are able to fit planes even for small specific objects and thus we can apply our approach in a wider variety of scenes.  Moreover the fitting procedure is made easier by the clustering and be done in real-time
compared to the 5 fps limitation in \cite{ardnt2020from}.

To avoid creating wrong planes that would corrupt the map, a plane is accepted if it is supported by enough inliers, which depends on the cluster class. For example, keyboards require at least 50 points to be inliers.

~\\
\textbf{Map optimization:} One way to add a structure constraint is to use a lagrangian multiplier \cite{nocedal2006numerical}, however this adds parameters that need to be estimated, thus we chose to include the constraint as a regularizer as can be seen in equation (\ref{eq:plane_BA}).

\begin{gather}
\label{eq:plane_BA}
\begin{aligned}
        \hat{\campose}, \hat{\point{w}} &= \argmin_{\campose, \point{w}}  \sum_{i,j}  \rho(||\imgpointij{i}{j}- proj(\camposetime{i}, \pointidx{w}{j})||_{\Sigma_{i,j}})  \\
        & + \sum_k\sum_{j \in O_k} \rho(||{\pi_{k}^\top} \pointidx{w}{j}||_{\sigma}) 
\end{aligned}
\raisetag{22pt}
\end{gather}
where $||\pi_{k}\pointidx{w}{j}||$ is the 3D distance between the 3D point $\pointidx{w}{j}$ (in homogeneous coordinates) and the plane $\pi_{k}$ which corresponds to the cluster $O_k$, $\sigma$ corresponds to its uncertainty, and $\rho$ is a robust cost function (in our case the Huber loss). 

Contrary to \cite{ardnt2020from} we do not project 3D points in their plane as some clusters may not be perfectly planar. However the strength of the constraint can be tuned by changing the value of $\sigma$. Moreover contrary to \cite{ardnt2020from} we do not optimize the planes, treating them as landmark but only use them as constraints on the map structure.



We optimize this equation using the framework g2o \cite{kummerle2011g}. We build a classical BA graph. Then we add an unary constraint to each point belonging to a planar cluster.  We consider those points to be outliers if their error is greater than the 95$^{th}$ percentile of a one-dimensional Chi-squared distribution. To account for points that would be outside of the local BA when a plane is fitted we chose to apply a global planar bundle adjustment after plane fitting. Furthermore to account for segmentation noise, we remove from the cluster points that are too far from the plane.

To optimize equation (\ref{eq:plane_BA}) we need to compute its jacobian. It is composed of the derivatives of the reprojection error with respect to camera poses and 3D points, which is the same as in classical BA \cite{dellaert2014visual}.
and of the derivatives of the point-plane error with respect to points position. This derivative can be computed as:
\begin{equation}
    \frac{\partial {\pi_{k}^\top} \pointidx{w}{j}}{\partial \pointidx{w}{j}} = {\pi_{k}^\top}
\end{equation}
\section{Experiments}
In this section we first present the implementation details of our approach, then we show on sequences from the TUM and the KITTI dataset \cite{sturm12iros, geiger2012we} the validity of our method.
\subsection{Implementation details}
S$^3$LAM runs at 20 fps and is based on the state of the art ORB-SLAM2 \cite{mur2017orb}. 
We do not take in account the time needed for the inference of the segmentation network as it depends on the choice of the model and on the GPU used. However we note that from detectron \cite{wu2019detectron2} model zoo, the inference time of panoptic segmentation networks varies from 53 to 98 ms with their setup, making it close to real time.
The optimization is done using the optimization framework g2o \cite{kummerle2011g}. 
All the experiments are performed on a desktop computer with an Intel Xeon @3.7GHz with 16 Gb of RAM and an Nvidia RTX2070. The value of the point-plane uncertainty $\sigma$ is set constant to $100$, which balances both errors as the value of the point-plane distance is around $10^{-2}$ m while the reprojection error is around one pixel. 
~\\
\textbf{Datasets.} Our approach is evaluated on sequences from the TUM RGB-D dataset \cite{sturm12iros} which provides a set of RGB frames associated to their groundtruth pose. We also show that our approach can work in larger scale and in outdoors scenes by evaluating it on sequences from the KITTI raw dataset \cite{geiger2012we}. This dataset contains sequences obtained using a camera mounted on a car in a wide variety of scenes. We evaluate on 4 sequences in which the road can be segmented and is planar and in which few objects are moving.
~\\
\textbf{Metrics.}
To account for the inherent stochasticity of ORB-SLAM2 we run each sequence 10 times and report the median of the RMSE of absolute trajectory error (ATE, defined in \cite{sturm12iros}) in the tables below. As our experiments are performed in a monocular setting, we scale and align the estimated trajectories with the ground truth as in \cite{mur2015orb}.

\subsection{Impact of the planes constraints on pose error }
In this section we study the impact of adding planar constraints to the classical BA formulation, using only planes robustly estimated from the 3D semantic map. We report the results of our experiments in table \ref{table:planar_BA}. 
We compare our approach to: the base system of S$^3$LAM, ORB-SLAM2 \cite{mur2017orb} and both the monocular and RGB-D approaches of Hosseinzadeh et al. \cite{hosseinzadeh2018structure, hosseinzadeh2019real} that report the greatest number of experiments and use both quadrics and planes. We also compare against the plane based approach of \cite{ardnt2020from} that report experiments on a few strongly planar sequences from the TUM dataset, as their code is not available we were not able to run experiments on new sequences. 

As we can see, planar constraints improve camera localization in most cases over ORB-SLAM 2. 

The most important improvement is obtained on almost perfectly planar scenes like fr1\_floor and fr3\_nostr\_text\_near which allow to easily fit planes. As we could expect using large planes gives better ATE improvements, the line fr3\_nostr\_text\_near (merged books) in table \ref{table:planar_BA} shows  our approach using a single plane for all book clusters, thus treating them as a single large cluster. An interesting way to improve our approach would thus to add plane-plane constraints to increase the spatial extent of small planes.
The sequences fr3\_nostr\_text\_near and fr3\_nostr\_text\_near (merged books) are run with the loop closure deactivated to better show the impact of our approach. The activation of loop closure is shown in the row fr3\_nostr\_text\_near (loop).
The sequences fr2\_xyz and fr2\_desk are the most challenging for our approach as the main planes corresponding to the table and the floor are not well segmented and cluttered, yielding to a noisy estimation.
Compared to the CNN based approach of \cite{hosseinzadeh2019real} we can see that we obtain better results for planar scenes, however in scenes in which planes are not well segmented, using quadrics brings a more important improvement, which can be related to the fact that object detection is less noisy than panoptic segmentation.
Compared to the approach of \cite{ardnt2020from} we obtain similar results, with almost equal means, however we argue that our approach is more generic as we are not limited to mostly planar scenes and we can run it in near real time.
We argue that those results and the reported means, show that our approach is generic as it improves camera pose estimation in a wide variety of scenes with a preference for planar scenes, while other approaches focus either on scenes containing objects or on perfectly planar scenes. 
To further demonstrate that our approach is generic we show in table \ref{table:kitti} the evaluation of our approach in a large scale scene, on the KITTI raw dataset, using the class \textit{road} as a plane. As we can see our approach works in large outdoors scenes and we improve camera pose estimation compared to ORB-SLAM2 \cite{mur2017orb}.

\begin{table*}[]
\centering
\caption{Comparison of the ATE (mm) of our our approach against state of the art on the TUM dataset.}
\resizebox{\textwidth}{!}{%
\begin{tabular}{@{}c||c||c|c|c|c|c@{}}
\toprule
Sequence &  \cite{hosseinzadeh2018structure} (RGB-D) &ORB-SLAM 2 \cite{mur2017orb} & \cite{hosseinzadeh2019real} (RGB) w. planes & \cite{hosseinzadeh2019real} (RGB) w. quadrics & \cite{ardnt2020from} & S$^3$LAM \\ \midrule
fr1\_xyz &9.6 &9.2 & 10.3 & 10.0 & - & \bf{8.8} \\
fr1\_floor & 13.8 &18.1 & 16.9 & - & - & \bf{14.7} \\
fr1\_desk &15.3 &13.9 & 12.9 & \bf{12.6} & - & 13.2 \\
fr2\_xyz &3.3 &2.4 & \bf{2.2} & \bf{2.2} & - & 2.4 \\
fr2\_desk & 12.0&8.0 & 7.3 & \bf{7.1} & - & 7.8 \\
fr3\_nost\_text\_near &- &20.3 & - & - & - & \bf{15.3} \\
fr3\_nost\_text\_near (merged books) &- &20.3 & -& - & - & \bf{13.5} \\
fr3\_nost\_text\_near (loop) &10.9 &14.5 & - & - & 11.4 & \bf{10.9} \\
fr3\_str\_text\_near & - & 14.0 & - & -& \bf{10.6} & 11.2 \\
fr3\_str\_text\_far& - & 10.6 & - & - & \bf{8.8} & 9.2 \\ \midrule
fr1 mean & 12.9& 13.9 & 13.4 & - & - & \bf{12.2} \\ 
fr2 mean &7.7 & 5.2 & 4.8 & \bf{4.7}  & - & 5.1 \\ 
fr3 mean & -& 13.0 & - & -  & \bf{10.3} & 10.4 \\ 

\bottomrule
\end{tabular}
}
\label{table:planar_BA}
\end{table*}

\begin{table}[]
\centering
\caption{Comparison of the ATE (cm) of our approach against ORB-SLAM2 on the Kitti raw dataset.}
\begin{tabular}{@{}c|c|c@{}}
\toprule
sequence & ORB-SLAM 2 \cite{mur2017orb}& \color{white} ----- \color{black} Ours \color{white} -----  \\ \midrule
0926-0011 & 17.7 &  \bf{15.5} \\
0926-0013 & 18.0 &    \bf{7.5} \\
0926-0014 & 76.2 &    \bf{64.5} \\
0926-0056 & 49.8 &    \bf{49.3} \\ \midrule
mean & 40.4 & \bf{34.2} \\ \bottomrule
\end{tabular}%
\label{table:kitti}

\end{table}

To further show that the estimated map structure is coherent we compute the angle product between pairs of normals that are supposed to be parallel at the end of a sequence (like pairs of books or books on a table for example). The minimum, maximum and median angles between normals are shown in table \ref{table:ps}. As we can see the estimated normals are pairwise coherent. Moreover the larger the planes are, the more points are extracted and the better the estimation is as visible in the last two lines of the table.

\begin{table}[]
\centering
\caption{Maximum, minimum and average values of angles between plane normals. The closer to 0 the better.}
\begin{tabular}{@{}c|c|c|c@{}}
 \toprule
Sequence & max. angle & min. angle & med. angle \\ \midrule
fr1\_desk & 3.6° & 2.4° & 2.9° \\
fr3\_nost\_text\_near & 1.4° & 0.8° & 0.8° \\
fr3\_nost\_text\_near (merged) & 0.0° & 0.0° & 0.0° \\ \bottomrule
\end{tabular}
\label{table:ps}
\end{table}

\subsection{Qualitative analysis of S$^3$LAM}
\begin{figure}[h!]
    \centering
    \includegraphics[width=\columnwidth]{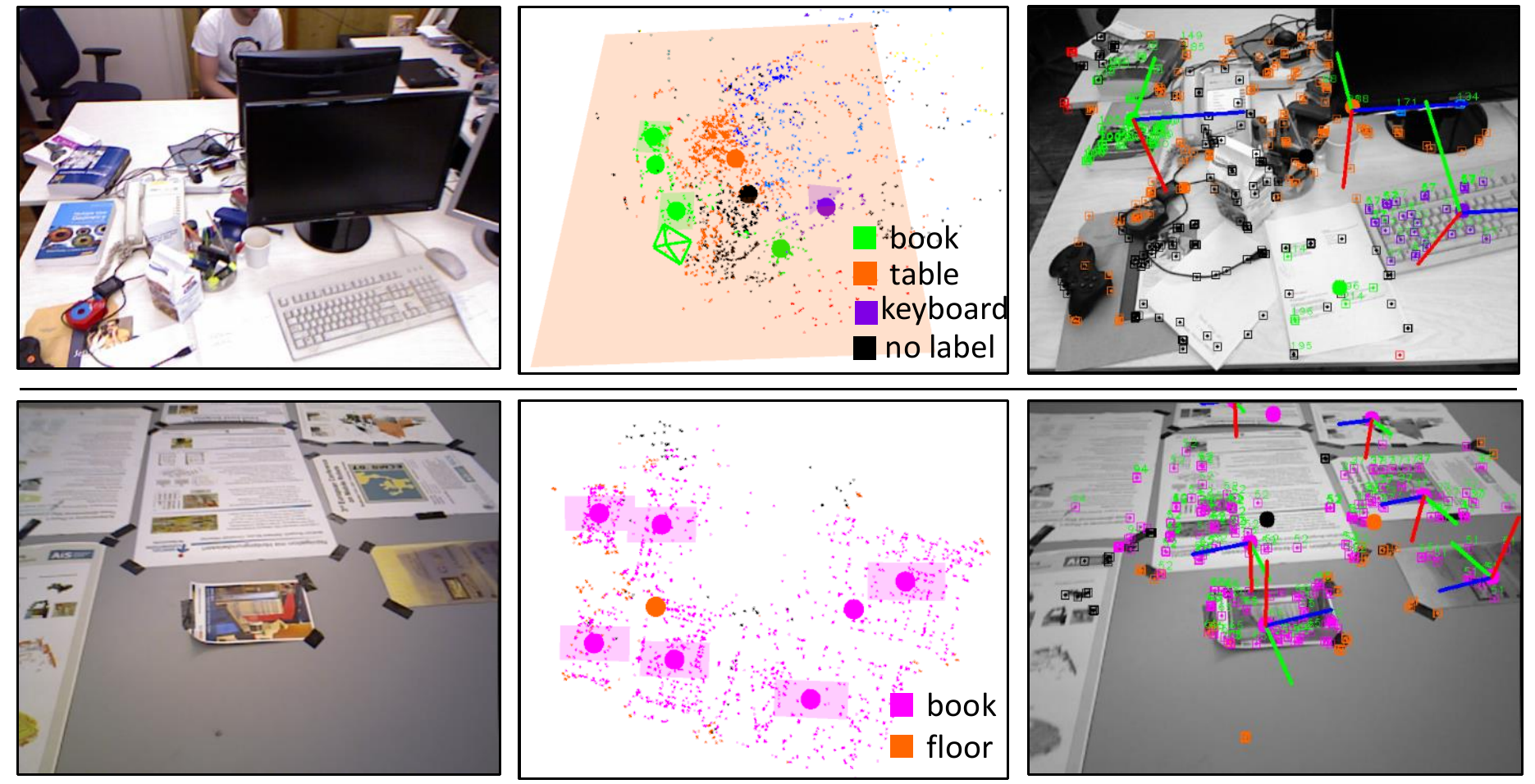}
    \caption{Examples of maps created by our approach for 2 sequences from \cite{sturm12iros}. From left to right: RGB input image from the sequence, S$^3$LAM map, frame with planes normals projected in red, in plane vectors are shown in blue and green. Keypoints are as well shown with a color corresponding to their class. }
    \label{fig:map}
\end{figure}

\begin{figure}[h!]
    \centering
    \includegraphics[width=\columnwidth]{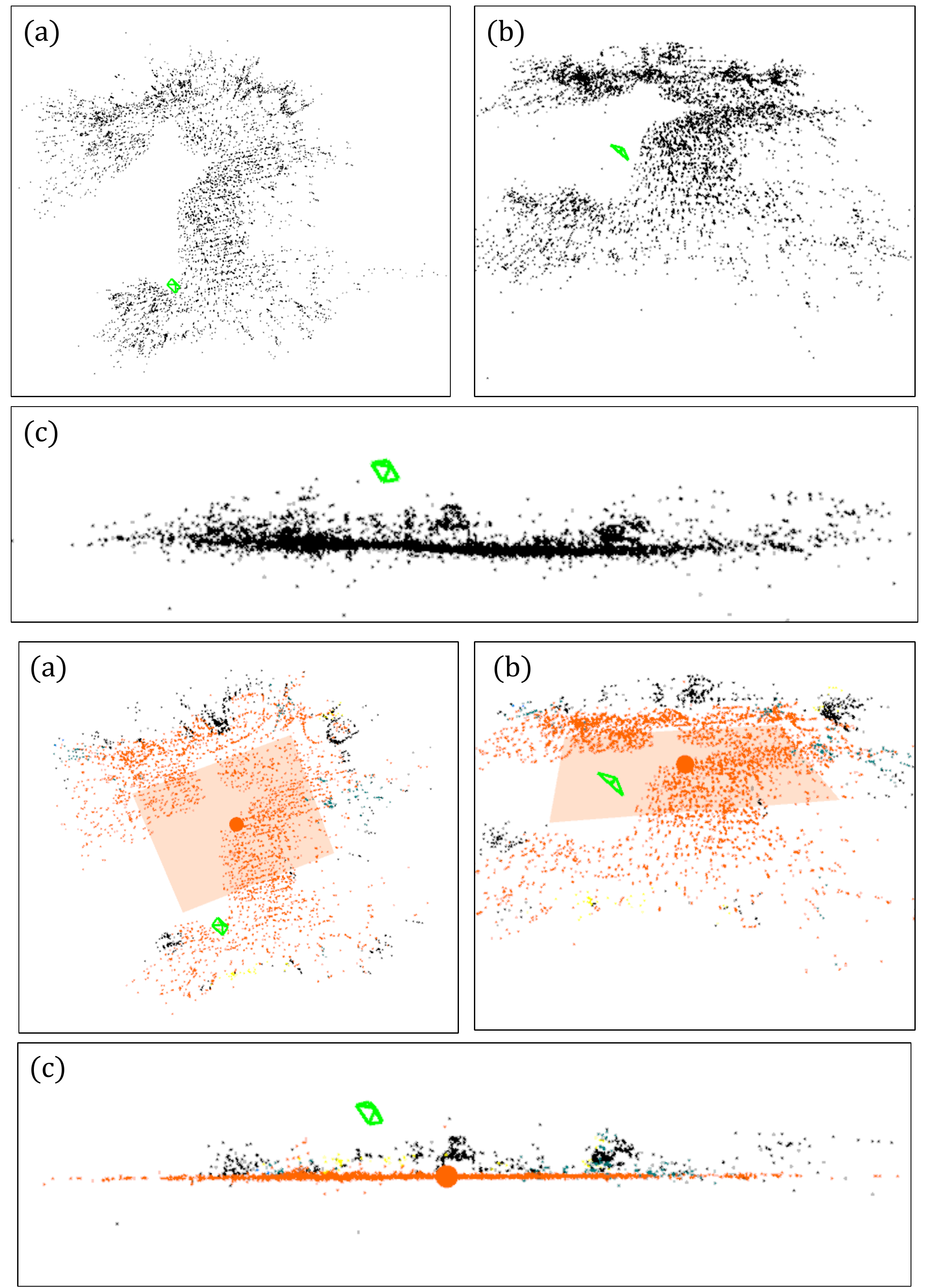}
    \caption{Examples of map created by ORB-SLAM2 \cite{mur2017orb} (top) and our approach (bottom) for the sequence fr1\_floor. (a) Map top view, (b) Map \nicefrac{3}{4} view, (c) Map side view. As we can see on the side view the floor is more planar with our approach. Black points that are outside of the plane are points that have not been segmented as "floor" and thus do not undergo the planar constraint.}
    \label{fig:comparison}
\end{figure}

We show in figure \ref{fig:map} some qualitative examples of maps obtained using our clustering approach compared to maps obtained using ORB-SLAM2 \cite{mur2017orb}. The goal of this figure is to show that the clusters and planes are well defined, to better see the effect of planes on the map quality we refer the reader to figure \ref{fig:comparison}. To construct these maps we used the following classes: table, keyboard, book. As we can see every object present in the scene has been uniquely clustered, leading to a more comprehensible and higher level map. We also show for each planar cluster the estimated planes, corresponding to the objects table, keyboard and book. Our approach yields a more physically accurate representation of the world, which can be easily used for augmented reality or robotics applications.  In figure \ref{fig:comparison} we show a comparison of the map obtained with ORB-SLAM2 and the map obtained using our planar BA. As we can see at the bottom, the lower part of the map corresponding to the floor (visible in orange in the bottom part of the map) is much more planar and coherent using our approach.

\section{Conclusion}
In this paper we proposed a new monocular semantic SLAM system called S$^3$LAM. 
Our method uses the 2D panoptic segmentation of a sequence of RGB images to create clusters of 3D points according to their class and instance. 
This clustering allows us to robustly estimate the structure of some clusters and modify the Bundle Adjutment formulation with structural constraints. We show on sequences from several public datasets that our approach lead to an improvement of camera pose estimation. 





\bibliographystyle{IEEEtran} 
\bibliography{IEEEabrv,IEEEexample}

\begin{thebibliography}{10}
\providecommand{\url}[1]{#1}
\csname url@rmstyle\endcsname
\providecommand{\newblock}{\relax}
\providecommand{\bibinfo}[2]{#2}
\providecommand\BIBentrySTDinterwordspacing{\spaceskip=0pt\relax}
\providecommand\BIBentryALTinterwordstretchfactor{4}
\providecommand\BIBentryALTinterwordspacing{\spaceskip=\fontdimen2\font plus
\BIBentryALTinterwordstretchfactor\fontdimen3\font minus
  \fontdimen4\font\relax}
\providecommand\BIBforeignlanguage[2]{{%
\expandafter\ifx\csname l@#1\endcsname\relax
\typeout{** WARNING: IEEEtran.bst: No hyphenation pattern has been}%
\typeout{** loaded for the language `#1'. Using the pattern for}%
\typeout{** the default language instead.}%
\else
\language=\csname l@#1\endcsname
\fi
#2}}

\bibitem{mur2017orb}
R.~Mur-Artal and J.~D. Tard{\'o}s, ``{ORB-SLAM}2: An open-source {SLAM} system
  for monocular, stereo, and {RGB-D} cameras,'' \emph{IEEE Trans. on Robotics},
  vol.~33, no.~5, pp. 1255--1262, 2017.

\bibitem{engel2014lsd}
J.~Engel, T.~Sch{\"o}ps, and D.~Cremers, ``{LSD-SLAM}: Large-scale direct
  monocular {SLAM},'' in \emph{European Conf. on computer vision}.\hskip 1em
  plus 0.5em minus 0.4em\relax Springer, 2014, pp. 834--849.

\bibitem{cadena2016past}
C.~Cadena, L.~Carlone, H.~Carrillo, Y.~Latif, D.~Scaramuzza, J.~Neira, I.~Reid,
  and J.~J. Leonard, ``Past, present, and future of simultaneous localization
  and mapping: Toward the robust-perception age,'' \emph{IEEE Trans. on
  robotics}, vol.~32, no.~6, pp. 1309--1332, 2016.

\bibitem{rosinol2020kimera}
A.~Rosinol, M.~Abate, Y.~Chang, and L.~Carlone, ``Kimera: an open-source
  library for real-time metric-semantic localization and mapping,'' in
  \emph{IEEE Int. Conf. on Robotics and Automation}.\hskip 1em plus 0.5em minus
  0.4em\relax IEEE, 2020, pp. 1689--1696.

\bibitem{runz2018maskfusion}
M.~Runz, M.~Buffier, and L.~Agapito, ``Maskfusion: Real-time recognition,
  tracking and reconstruction of multiple moving objects,'' in \emph{2018 IEEE
  International Symposium on Mixed and Augmented Reality (ISMAR)}, 2018, pp.
  10--20.

\bibitem{redmon2016you}
J.~Redmon, S.~Divvala, R.~Girshick, and A.~Farhadi, ``You only look once:
  Unified, real-time object detection,'' in \emph{IEEE Conf. on computer vision
  and pattern recognition}, 2016, pp. 779--788.

\bibitem{he2017mask}
K.~He, G.~Gkioxari, P.~Doll{\'a}r, and R.~Girshick, ``Mask {R}-{CNN},'' in
  \emph{IEEE Int. Conf. on computer vision}, 2017, pp. 2961--2969.

\bibitem{kirillov2019panoptic}
A.~Kirillov, R.~Girshick, K.~He, and P.~Doll{\'a}r, ``Panoptic feature pyramid
  networks,'' in \emph{IEEE/CVF Conf. on Computer Vision and Pattern
  Recognition}, 2019, pp. 6399--6408.

\bibitem{tateno2017cnn}
K.~Tateno, F.~Tombari, I.~Laina, and N.~Navab, ``{CNN}-{SLAM}: Real-time dense
  monocular {SLAM} with learned depth prediction,'' in \emph{IEEE Conf. on
  Computer Vision and Pattern Recognition}, 2017, pp. 6243--6252.

\bibitem{mccormac2017semanticfusion}
J.~McCormac, A.~Handa, A.~Davison, and S.~Leutenegger, ``Semanticfusion: Dense
  3{D} semantic mapping with convolutional neural networks,'' in \emph{2017
  IEEE Int. Conf. on Robotics and automation (ICRA)}, 2017, pp. 4628--4635.

\bibitem{schonberger2018semantic}
J.~L. Sch{\"o}nberger, M.~Pollefeys, A.~Geiger, and T.~Sattler, ``Semantic
  visual localization,'' in \emph{IEEE Conf. on Computer Vision and Pattern
  Recognition}, 2018, pp. 6896--6906.

\bibitem{toft2017long}
C.~Toft, C.~Olsson, and F.~Kahl, ``Long-term 3{D} localization and pose from
  semantic labellings,'' in \emph{{IEEE} Int. Conf. on Computer Vision
  Workshops}, 2017, pp. 650--659.

\bibitem{toft2018semantic}
C.~Toft, E.~Stenborg, L.~Hammarstrand, L.~Brynte, M.~Pollefeys, T.~Sattler, and
  F.~Kahl, ``Semantic match consistency for long-term visual localization,'' in
  \emph{European Conf. on Computer Vision (ECCV)}, 2018, pp. 383--399.

\bibitem{bescos2018dynaslam}
B.~Bescos, J.~M. F{\'a}cil, J.~Civera, and J.~Neira, ``Dyna{SLAM}: Tracking,
  mapping, and inpainting in dynamic scenes,'' \emph{{IEEE} Robotics and
  Automation Letters}, vol.~3, no.~4, pp. 4076--4083, 2018.

\bibitem{huang2020clustervo}
J.~Huang, S.~Yang, T.-J. Mu, and S.-M. Hu, ``Cluster{VO}: Clustering moving
  instances and estimating visual odometry for self and surroundings,'' in
  \emph{IEEE/CVF Conf. on Computer Vision and Pattern Recognition}, 2020, pp.
  2168--2177.

\bibitem{yang2019cubeslam}
S.~Yang and S.~Scherer, ``Cube{SLAM}: Monocular 3-d object {SLAM},'' \emph{IEEE
  Trans. on Robotics}, vol.~35, no.~4, pp. 925--938, 2019.

\bibitem{salas2013slam++}
R.~F. Salas-Moreno, R.~A. Newcombe, H.~Strasdat, P.~H. Kelly, and A.~J.
  Davison, ``{SLAM}++: Simultaneous localisation and mapping at the level of
  objects,'' in \emph{IEEE Conf. on computer vision and pattern recognition},
  2013, pp. 1352--1359.

\bibitem{hosseinzadeh2019real}
M.~Hosseinzadeh, K.~Li, Y.~Latif, and I.~Reid, ``Real-time monocular
  object-model aware sparse {SLAM},'' in \emph{2019 Int. Conf. on Robotics and
  Automation (ICRA)}, 2019, pp. 7123--7129.

\bibitem{nicholson2018quadricslam}
L.~Nicholson, M.~Milford, and N.~S{\"u}nderhauf, ``Quadric{SLAM}: Dual quadrics
  from object detections as landmarks in object-oriented {SLAM},'' \emph{IEEE
  Robotics and Automation Letters}, vol.~4, no.~1, pp. 1--8, 2018.

\bibitem{galvez2016real}
D.~G{\'a}lvez-L{\'o}pez, M.~Salas, J.~D. Tard{\'o}s, and J.~Montiel,
  ``Real-time monocular object {SLAM},'' \emph{Robotics and Autonomous
  Systems}, vol.~75, pp. 435--449, 2016.

\bibitem{fioraio2013joint}
N.~Fioraio and L.~Di~Stefano, ``Joint detection, tracking and mapping by
  semantic bundle adjustment,'' in \emph{IEEE Conf. on Computer Vision and
  Pattern Recognition}, 2013, pp. 1538--1545.

\bibitem{civera2011towards}
J.~Civera, D.~G{\'a}lvez-L{\'o}pez, L.~Riazuelo, J.~D. Tard{\'o}s, and J.~M.~M.
  Montiel, ``Towards semantic {SLAM} using a monocular camera,'' in \emph{2011
  IEEE/RSJ Int. Conf. on Intelligent Robots and Systems}, 2011, pp. 1277--1284.

\bibitem{rad2017bb8}
M.~Rad and V.~Lepetit, ``{BB}8: A scalable, accurate, robust to partial
  occlusion method for predicting the 3{D} poses of challenging objects without
  using depth,'' in \emph{IEEE Int. Conf. on Computer Vision}, 2017, pp.
  3828--3836.

\bibitem{gonzalez2021l6dnet}
M.~Gonzalez, A.~Kacete, A.~Murienne, and E.~Marchand, ``L6dnet: Light 6 {DoF}
  network for robust and precise object pose estimation with small datasets,''
  \emph{IEEE Robotics and Automation Letters}, 2021.

\bibitem{xiang2017posecnn}
Y.~Xiang, T.~Schmidt, V.~Narayanan, and D.~Fox, ``Pose{CNN}: A convolutional
  neural network for 6d object pose estimation in cluttered scenes,''
  \emph{arXiv preprint arXiv:1711.00199}, 2017.

\bibitem{kaess2015simultaneous}
M.~Kaess, ``Simultaneous localization and mapping with infinite planes,'' in
  \emph{2015 IEEE Int. Conf. on Robotics and Automation (ICRA)}, 2015, pp.
  4605--4611.

\bibitem{ardnt2020from}
C.~{Arndt}, R.~{Sabzevari}, and J.~{Civera}, ``From points to planes - adding
  planar constraints to monocular {SLAM} factor graphs,'' in \emph{2020
  IEEE/RSJ Int. Conf. on Intelligent Robots and Systems (IROS)}, 2020, pp.
  4917--4922.

\bibitem{yang2016pop}
S.~Yang, Y.~Song, M.~Kaess, and S.~Scherer, ``Pop-up {SLAM}: Semantic monocular
  plane {SLAM} for low-texture environments,'' in \emph{2016 IEEE/RSJ Int.
  Conf. on Intelligent Robots and Systems (IROS)}, 2016, pp. 1222--1229.

\bibitem{wu2019detectron2}
Y.~Wu, A.~Kirillov, F.~Massa, W.-Y. Lo, and R.~Girshick, ``Detectron2,''
  \url{https://github.com/facebookresearch/detectron2}, 2019.

\bibitem{rublee2011orb}
E.~Rublee, V.~Rabaud, K.~Konolige, and G.~Bradski, ``{ORB}: An efficient
  alternative to {SIFT} or {SURF},'' in \emph{2011 Int. Conf. on computer
  vision}, 2011, pp. 2564--2571.

\bibitem{mouragnon2006real}
E.~Mouragnon, M.~Lhuillier, M.~Dhome, F.~Dekeyser, and P.~Sayd, ``Real time
  localization and 3{D} reconstruction,'' in \emph{2006 IEEE Computer Society
  Conf. on Computer Vision and Pattern Recognition (CVPR'06)}, vol.~1, 2006,
  pp. 363--370.

\bibitem{klein2007parallel}
G.~Klein and D.~Murray, ``Parallel tracking and mapping for small {AR}
  workspaces,'' in \emph{2007 6th IEEE and ACM international symposium on mixed
  and augmented reality}, 2007, pp. 225--234.

\bibitem{triggs1999bundle}
B.~Triggs, P.~F. McLauchlan, R.~I. Hartley, and A.~W. Fitzgibbon, ``Bundle
  adjustment—a modern synthesis,'' in \emph{International workshop on vision
  algorithms}.\hskip 1em plus 0.5em minus 0.4em\relax Springer, 1999, pp.
  298--372.

\bibitem{sunderhauf2017meaningful}
N.~S{\"u}nderhauf, T.~T. Pham, Y.~Latif, M.~Milford, and I.~Reid, ``Meaningful
  maps with object-oriented semantic mapping,'' in \emph{2017 IEEE/RSJ Int.
  Conf. on Intelligent Robots and Systems (IROS)}, 2017, pp. 5079--5085.

\bibitem{yang2019monocular}
S.~Yang and S.~Scherer, ``Monocular object and plane {SLAM} in structured
  environments,'' \emph{IEEE Robotics and Automation Letters}, vol.~4, no.~4,
  pp. 3145--3152, 2019.

\bibitem{hsiao2017keyframe}
M.~Hsiao, E.~Westman, G.~Zhang, and M.~Kaess, ``Keyframe-based dense planar
  {SLAM},'' in \emph{2017 IEEE Int. Conf. on Robotics and Automation (ICRA)},
  2017, pp. 5110--5117.

\bibitem{hosseinzadeh2018structure}
M.~Hosseinzadeh, Y.~Latif, T.~Pham, N.~Suenderhauf, and I.~Reid, ``Structure
  aware slam using quadrics and planes,'' in \emph{Asian Conf. on Computer
  Vision}.\hskip 1em plus 0.5em minus 0.4em\relax Springer, 2018, pp. 410--426.

\bibitem{nocedal2006numerical}
J.~Nocedal and S.~Wright, \emph{Numerical optimization}.\hskip 1em plus 0.5em
  minus 0.4em\relax Springer Science \& Business Media, 2006.

\bibitem{kummerle2011g}
R.~K{\"u}mmerle, G.~Grisetti, H.~Strasdat, K.~Konolige, and W.~Burgard, ``g2o:
  A general framework for graph optimization,'' in \emph{2011 IEEE Int. Conf.
  on Robotics and Automation}, 2011, pp. 3607--3613.

\bibitem{dellaert2014visual}
F.~Dellaert, ``Visual slam tutorial: Bundle adjustment,'' \emph{CVPR'14
  tutorial}, 2014.

\bibitem{sturm12iros}
J.~Sturm, N.~Engelhard, F.~Endres, W.~Burgard, and D.~Cremers, ``A benchmark
  for the evaluation of {RGB-D SLAM} systems,'' in \emph{Proc. of the Int.
  Conf. on Intelligent Robot Systems (IROS)}, Oct. 2012.

\bibitem{geiger2012we}
A.~Geiger, P.~Lenz, and R.~Urtasun, ``Are we ready for autonomous driving? the
  kitti vision benchmark suite,'' in \emph{2012 IEEE Conf. on computer vision
  and pattern recognition}, 2012, pp. 3354--3361.

\bibitem{mur2015orb}
R.~Mur-Artal, J.~M.~M. Montiel, and J.~D. Tardos, ``{ORB-SLAM}: a versatile and
  accurate monocular {SLAM} system,'' \emph{IEEE Trans. on robotics}, vol.~31,
  no.~5, pp. 1147--1163, 2015.

\end{thebibliography}
\addtolength{\textheight}{-12cm} 
\end{document}